\documentclass[pdflatex,sn-mathphys-num]{sn-jnl}% Math and Physical Sciences Numbered Reference Style
\usepackage{graphicx}%
\usepackage{multirow}%
\usepackage{amsmath,amssymb,amsfonts}%
\usepackage{amsthm}%
\usepackage{mathrsfs}%
\usepackage[title]{appendix}%
\usepackage{xcolor}%
\usepackage{textcomp}%
\usepackage{manyfoot}%
\usepackage{booktabs}%
\usepackage{algorithm}%
\usepackage{algorithmicx}%
\usepackage{algpseudocode}%
\usepackage{listings}%
\theoremstyle{thmstyleone}%
\theoremstyle{thmstyletwo}%

\theoremstyle{thmstylethree}%

\begin{document}

\title[Article Title]{MS2MetGAN: Latent-space adversarial training for metabolite–spectrum matching in MS/MS database search}


\author[1]{\fnm{Meng} \sur{Tsai}}\email{wmf223@mocs.utc.edu}

\author[1]{\fnm{Alexzander} \sur{Dwyer}}\email{wpx739@mocs.utc.edu}

\author[2]{\fnm{Estelle} \sur{Nuckels}}\email{estelle.nuckels@mga.edu}

\author*[1]{\fnm{Yingfeng} \sur{Wang}}\email{yingfeng-wang@utc.edu}

\affil[1]{\orgdiv{Department of Computer Science and Engineering}, \orgname{University of Tennessee at Chattanooga}, \orgaddress{\street{615 McCallie Ave}, \city{Chattanooga}, \postcode{37403}, \state{Tennessee}, \country{United States}}}

\affil[2]{\orgdiv{Department of Natural Sciences}, \orgname{Middle Georgia State University}, \orgaddress{\street{100 University Pkwy}, \city{Macon}, \postcode{31206}, \state{Georgia}, \country{United States}}}


\abstract{Database search is a widely used approach for identifying metabolites from tandem mass spectra (MS/MS). In this strategy, an experimental spectrum is matched against a user-specified database of candidate metabolites, and candidates are ranked such that true metabolite–spectrum matches receive the highest scores. Machine-learning methods have been widely incorporated into database-search–based identification tools and have substantially improved performance. To further improve identification accuracy, we propose a new framework for generating negative training samples. The framework first uses autoencoders to learn latent representations of metabolite structures and MS/MS spectra, thereby recasting metabolite–spectrum matching as matching between latent vectors. It then uses a GAN to generate latent vectors of decoy metabolites and constructs decoy metabolite–spectrum matches as negative samples for training. Experimental results show that our tool, MS2MetGAN, achieves better overall performance than existing metabolite identification methods.}

\keywords{Metabolite, Tandem mass spectrum, Autoencoder, Generative adversarial network}



\maketitle

\section{Introduction}\label{sec1}

Metabolites (molecules associated with metabolic function, with molecular masses typically ranging from 50 to 1500 Da) are of significant biological and chemical importance. Whether they are amino acids, sugars, lipids, or drug byproducts, the potential of rapid metabolite identification to transform biochemistry \cite{Giera2022, Manickam2023, Elshafie2023}, drug discovery \cite{Baker2023, Qiu2023, Castelli2022} and the practice of medicine \cite{Qiu2023, Castelli2022, Nwakoby2025} has been recognized for decades \cite{Clarke2021}. However, the sheer number and diversity of metabolites make comprehensive identification both vitally important and extremely daunting. In untargeted metabolite identification, where metabolites are not pre-selected, liquid chromatography–electrospray ionization–tandem mass spectrometry (LC–ESI–MS/MS) is a preferred technique as it separates metabolites by LC, ionizes them by ESI, and measures them by MS/MS, offering high sensitivity, straightforward sample preparation, and highly informative measurements \cite{Verdegem2016}. 

An MS/MS spectrum comprises the mass-to-charge (m/z) ratios and corresponding intensities of a metabolite and its fragment ions. However, computational identification of metabolites from MS/MS data remains challenging. Most existing tools adopt a database-search strategy: given an experimental MS/MS spectrum, they query a user-specified database of metabolite structures and rank candidate metabolites by how well they match the spectrum. Many database-search methods incorporate machine-learning models to improve candidate ranking \cite{Verdegem2016, Ruttkies2019, Dührkop2015, Li2020, Wang2021, Cao2021, Chen2024, Xie2025}. Most of these models require negative samples during training, which are commonly constructed using mismatched compounds \cite{Verdegem2016, Ruttkies2019, Dührkop2015} or spectra \cite{Li2020, Chen2024, Xie2025}.

%MAGMa+ \cite{Verdegem2016}: the negative samples are real metabolites from the Metlin-derived dataset that belong to the other class
%MetFrag \cite{Ruttkies2019}: competing candidate compounds
%CSI:FingerID \cite{Dührkop2015}: the negative samples are training compounds that lack the specific fingerprint property being learned
%SF-Matching \cite{Li2020}: train a binary classifier for each subfragment, negative samples. spectra that do not contain a peak corresponding to that fragment
%CFM-ID \cite{Wang2021}: not contrastive, does not use negative samples in the training.
%MolDiscovery \cite{Cao2021}: does not use negative samples in the training.
%CMSSP \cite{Chen2024}: other compounds, but does not mention isomers
%CSU-MS\textsuperscript{2} \cite{Xie2025}: other compounds, e.g., isomers.

When machine learning is used to distinguish true metabolite–spectrum matches (MSMs) from decoy MSMs, the quality of negative training samples is crucial. Unfortunately, training is often constrained by the available decoy metabolites, whose structures are not sufficiently similar to those of the true metabolites. In this study, we propose a new method to address this limitation in MS2MetGAN, a database-search tool for identifying metabolites from MS/MS data. Our method uses autoencoders to embed metabolite structures and MS/MS spectra into latent numerical vectors, enabling MSMs to be represented as pairs of latent vectors (hereafter, latent MSMs). This method further introduces a generative adversarial network (GAN) with two components: a generator and a discriminator. The generator learns to produce synthetic latent structure vectors conditioned on spectrum latent vectors, while the discriminator learns to distinguish true latent MSMs from synthetic latent MSMs formed by pairing generated structure codes with true spectrum codes. The resulting optimized discriminator serves as the classifier for metabolite identification from MS/MS data. This design reframes decoy generation from synthesizing decoy compound structures to producing decoy latent vectors that represent compound structures, substantially reducing the difficulty of generating decoys.

\section{Results}\label{sec2}
\subsection{Overview}\label{subsec2-1}

\begin{figure}[]
\centering
\includegraphics[width=\linewidth]{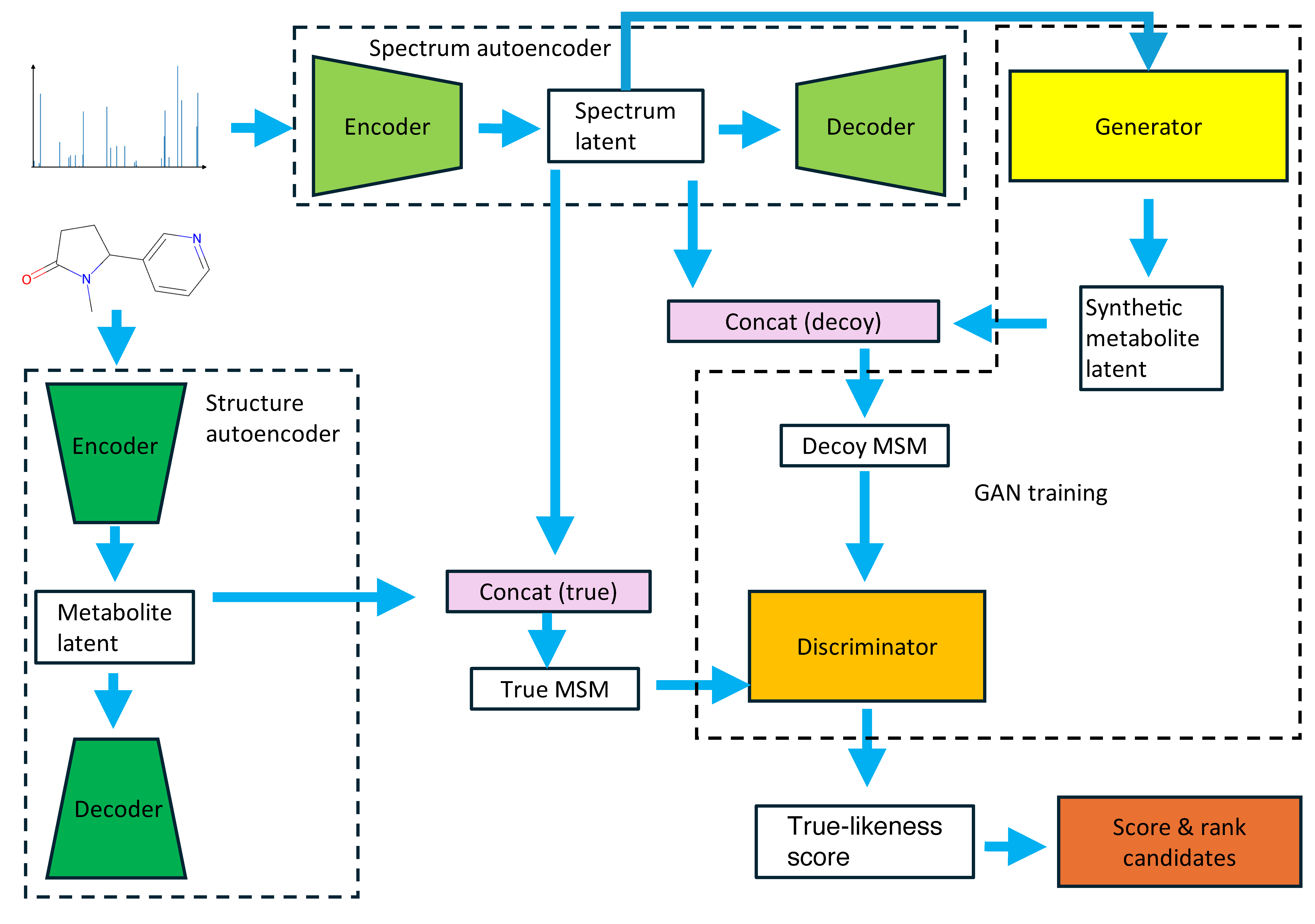}
\caption{Overview of the method. MS/MS spectra and metabolite structures are encoded into latent spectrum vectors and latent metabolite vectors by a spectrum autoencoder and a structure autoencoder, respectively. The generative adversarial network (GAN) comprises a generator and a discriminator. The generator produces synthetic latent metabolite vectors conditioned on latent spectrum vectors, whereas the discriminator assigns a true-likeness score to each latent metabolite–spectrum match (MSM), distinguishing true MSMs from decoy MSMs formed by pairing synthetic metabolite latents with the corresponding spectrum latents. These scores are then used to rank metabolite candidates.}\label{fig_1}
\end{figure}

MS2MetGAN combines autoencoders and a GAN to train a classifier that distinguishes true latent MSMs from false latent MSMs. An overview of the method is shown in Fig.~\ref{fig_1}. This method uses a graph-transformer neural network to encode molecular structures into latent vectors. Similarly, it uses an MLP autoencoder to encode MS/MS spectra into latent vectors. MS2MetGAN then concatenates the two latent vectors to form a latent MSM representation, which is fed into an MLP classifier to predict whether the match is true. To further strengthen the classifier, this method introduces a GAN. The GAN generator is an MLP that maps a spectrum latent code to a molecular-structure latent code, and the MSM classifier serves as the discriminator. Conditioned on true spectrum latent codes, the generator produces structure latent codes. Pairing these generated structure codes with the corresponding true spectrum codes yields challenging decoy MSMs that are used to train and improve the discriminator (that is, the classifier). The trained discriminator scores latent MSMs to be used within a database-search framework for metabolite identification via MS/MS data.

%\begin{verbatim}
%\end{verbatim}

\subsection{Identifying metabolites from MS/MS spectra}\label{subsec_experiments}
\subsubsection{Experimental setting}\label{subsubsec_setting}
This study constructs a training dataset for MS2MetGAN (see the ``Training dataset'' section in Supplementary Information for more details). Positive training samples consist of standard MS/MS spectra acquired in positive ion mode ([M+H]\textsuperscript{+}) and their associated compounds, obtained from the MassBank of North America (MoNA). The dataset also includes negative training samples for the initial training of the proposed discriminator: the spectra are identical to those in the positive samples, while the compounds are structural isomers of the corresponding positive-sample compounds retrieved from PubChem \cite{Kim2025}. 

This study also constructs test datasets to compare MS2MetGAN with competing metabolite identification tools (see the ``Test datasets'' section in Supplementary Information for more details). We evaluate performance on nine benchmark sets of standard MS/MS spectra acquired in positive ion mode ([M+H]\textsuperscript{+}). Five benchmarks are derived from the Critical Assessment of Small Molecule Identification (CASMI) contests: CASMI2016FP, CASMI2017FP, CASMI2016SP, CASMI2017SP, and CASMI2022P. CASMI2016FP and CASMI2017FP include all [M+H]\textsuperscript{+} standard spectra from CASMI 2016 and CASMI 2017, respectively, whereas CASMI2016SP and CASMI2017SP correspond to the subsets used by SF-Matching \cite{Li2020}. CASMI2022P is used in CSU-MS2 \cite{Xie2025}. For this benchmark, we remove spectra whose associated compounds appear in the training dataset. We also evaluate on two positive-ion datasets from GNPS \cite{Wang2016}, GNPS-S and GNPS-M, which are used by SF-Matching \cite{Li2020} and MolDiscovery \cite{Cao2021}, respectively. Finally, we include the EMBL-MCF and MoNA benchmarks, which are also employed by SF-Matching \cite{Li2020} and MolDiscovery \cite{Cao2021}, respectively. For each benchmark set, the precursor information and MS/MS peak list of each test spectrum are searched against two compound databases. The first database comprises all compounds in MetaCyc (v29.0) together with the true compounds in the benchmark set. The second database consists of the true compounds and their isomer decoys. For each spectrum, PubChem is queried to retrieve isomeric compounds with the same chemical formula as the corresponding true compound. For either database, candidate molecules are defined as those with masses within 0.02 Da of the spectrum’s precursor mass.

This study compares performances of MS2MetGAN to that of eight competing metabolite identification tools with their default parameter settings on benchmark sets: MIDAS \cite{Wang2014}, MAGMa+ \cite{Verdegem2016}, MetFrag \cite{Ruttkies2019}, CSI:FingerID \cite{Dührkop2015}, SF-Matching \cite{Li2020}, CFM-ID \cite{Wang2021}, CMSSP \cite{Chen2024}, and CSU-MS2 \cite{Xie2025} (see the ``Baseline tools'' section in Supplementary Information for more details). It is worth noting that we do not include MolDiscovery \cite{Cao2021} in the evaluation because it does not currently provide an implementation that supports batch processing.  

\subsubsection{Metabolite Identification Performance}

The metabolite identification performance of the baseline tools and MS2MetGAN against the MetaCyc database is reported in Supplementary Tables S1–S9 and summarized in Table~\ref{tab:MetaCyc29-All-Models}. MS2MetGAN achieves an average accuracy more than seven percentage points higher than the second-best tool, MIDAS, and attains the highest accuracy on four benchmark sets. Supplementary Tables S10–S18 report performance against isomer decoys, with results summarized in Table~\ref{tab:PubChem-All-Models}. Under this setting, MS2MetGAN achieves an average accuracy more than ten percentage points higher than the second-best tool, SF-Matching, and attains the highest accuracy on five benchmark sets. Table~\ref{tab:Baseline-against-MS2MetGAN} reports pairwise comparisons between MS2MetGAN and the nine baseline tools. MS2MetGAN outperforms every baseline tool on a larger fraction of test datasets when searching against MetaCyc, and likewise when searching against isomer decoys.

The discriminator is also evaluated at each GAN training round (see the ``GAN architecture and training'' section for details). GAN-0 denotes the discriminator from the initial round, whereas GAN-9 denotes the discriminator from the ninth round, which corresponds to MS2MetGAN. Supplementary Tables S19–S27 report the performance of all discriminators except GAN-9 (reported in Supplementary Table S9) when searching against MetaCyc. Table~\ref{tab:MetaCyc-GANs} summarizes these results. GAN-9 (MS2MetGAN) achieves the highest average accuracy. Moreover, the standard deviations of accuracies for GAN-1 through GAN-9 are consistently much lower than that of GAN-0, suggesting that the GAN training procedure improves the stability of performance across test datasets. Similarly, Supplementary Tables S28–S36 report the performance of all discriminators except GAN-9, which is reported in Supplementary Table S18, when searching against isomer decoys. Table~\ref{tab:PubChem-GANs} summarizes these results. GAN-9 (MS2MetGAN) again achieves the highest average accuracy, and the standard deviations of accuracies for GAN-1 through GAN-9 remain consistently much lower than that of GAN-0.

\begin{table}[]
\caption{Identification accuracies of baseline tools and MS2MetGAN for database searching against MetaCyc (bold indicates the highest accuracy)} 

\label{tab:MetaCyc29-All-Models}
\begin{tabular}{lccccc}
\hline
            & MIDAS   & SF-Matching      & MAGMa+  & CSI:FingerID     & CFM-ID  \\
\hline
CASMI2016SP & 90.98\% & \textbf{99.18\%} & 54.92\% & 59.84\%          & 69.67\% \\
CASMI2017SP & 88.10\% & 83.33\%          & 59.52\% & 64.29\%          & 61.90\% \\
CASMI2016FP & 85.04\% & \textbf{98.16\%} & 57.22\% & 86.88\%          & 66.40\% \\
CASMI2017FP & 67.12\% & 61.64\%          & 51.37\% & 30.82\%          & 42.47\% \\
CASMI2022P  & 68.94\% & 58.39\%          & 54.04\% & \textbf{77.64\%} & 65.84\% \\
GNPS-S      & 54.92\% & 44.04\%          & 30.57\% & 34.72\%          & 8.29\%  \\
GNPS-M      & 73.20\% & \textbf{92.27\%} & 53.61\% & 22.68\%          & 29.90\% \\
EMBL-MCF    & 53.38\% & 20.90\%          & 20.90\% & 50.16\%          & 8.36\%  \\
MoNA        & 41.23\% & 34.21\%          & 32.46\% & 19.88\%          & 6.73\%  \\
Mean        & 69.21\% & 65.79\%          & 46.07\% & 49.66\%          & 39.95\% \\
\hline
            & MetFrag & CMSSP            & CSU-MS2 & MS2MetGAN        &         \\
\hline
CASMI2016SP & 42.62\% & 88.52\%          & 26.23\% & 81.97\%          &         \\
CASMI2017SP & 26.19\% & \textbf{92.86\%} & 54.76\% & 88.10\%          &         \\
CASMI2016FP & 33.86\% & 85.04\%          & 24.41\% & 85.56\%          &         \\
CASMI2017FP & 17.81\% & 75.34\%          & 30.14\% & \textbf{82.19\%} &         \\
CASMI2022P  & 65.22\% & 44.72\%          & 41.61\% & 37.89\%          &         \\
GNPS-S      & 6.74\%  & 18.65\%          & 43.01\% & \textbf{73.06\%} &         \\
GNPS-M      & 17.01\% & 85.57\%          & 44.33\% & 72.16\%          &         \\
EMBL-MCF    & 2.57\%  & 11.25\%          & 22.19\% & \textbf{93.25\%} &         \\
MoNA        & 3.80\%  & 6.73\%           & 57.02\% & \textbf{72.81\%} &         \\
Mean        & 23.98\% & 56.52\%          & 38.19\% & \textbf{76.33\%} &         \\
\hline
\end{tabular}
\end{table}

\begin{table}[]
\caption{Identification accuracies of baseline tools and MS2MetGAN for database searching against isomer decoys (bold indicates the highest accuracy)}
\label{tab:PubChem-All-Models}
\begin{tabular}{lccccc}
\hline
            & MIDAS            & SF-Matching      & MAGMa+  & CSI:FingerID     & CFM-ID  \\
\hline
CASMI2016SP & 54.92\%          & \textbf{98.36\%} & 32.79\% & 58.20\%          & 61.48\% \\
CASMI2017SP & 57.14\%          & 80.95\%          & 30.95\% & 69.05\%          & 45.24\% \\
CASMI2016FP & 54.59\%          & \textbf{95.28\%} & 33.07\% & 83.99\%          & 57.48\% \\
CASMI2017FP & 39.73\%          & 60.27\%          & 27.40\% & 31.51\%          & 42.47\% \\
CASMI2022P  & \textbf{42.86\%} & 34.16\%          & 29.81\% & 29.19\%          & 34.78\% \\
GNPS-S      & 40.42\%          & 64.77\%          & 16.58\% & 34.72\%          & 72.02\% \\
GNPS-M      & 47.94\%          & \textbf{93.81\%} & 30.41\% & 21.65\%          & 38.14\% \\
EMBL-MCF    & 43.41\%          & 40.84\%          & 13.18\% & 65.60\%          & 54.02\% \\
MoNA        & 26.02\%          & 50.00\%          & 22.22\% & 18.13\%          & 62.57\% \\
Mean        & 45.23\%          & 68.72\%          & 26.27\% & 45.78\%          & 52.02\% \\
\hline
            & MetFrag          & CMSSP            & CSU-MS2 & MS2MetGAN        &         \\
\hline
CASMI2016SP & 51.64\%          & 89.34\%          & 7.38\%  & 86.07\%          &         \\
CASMI2017SP & 30.95\%          & 59.52\%          & 16.67\% & \textbf{90.48\%} &         \\
CASMI2016FP & 50.92\%          & 89.76\%          & 7.61\%  & 87.40\%          &         \\
CASMI2017FP & 48.63\%          & 80.14\%          & 20.55\% & \textbf{86.30\%} &         \\
CASMI2022P  & 38.51\%          & 24.22\%          & 11.80\% & 37.89\%          &         \\
GNPS-S      & 21.24\%          & 23.83\%          & 29.02\% & \textbf{75.65\%} &         \\
GNPS-M      & 44.85\%          & 79.38\%          & 28.87\% & 79.90\%          &         \\
EMBL-MCF    & 21.54\%          & 25.08\%          & 22.19\% & \textbf{93.25\%} &         \\
MoNA        & 26.02\%          & 8.48\%           & 57.31\% & \textbf{77.19\%} &         \\
Mean        & 37.14\%          & 53.31\%          & 22.38\% & \textbf{79.35\%} &         \\
\hline
\end{tabular}
%\footnotetext[1]{Bold indicates the best performance.}
\end{table}

\begin{table}[]
\caption{Percentage of test datasets on which MS2MetGAN outperforms baseline tools in pairwise comparisons}
\label{tab:Baseline-against-MS2MetGAN}
\begin{tabular}{lcccc}
\hline
                 & MIDAS    & SF-Matching & MAGMa+   & CSI:FingerID \\
\hline
MetaCyc Database & 66.67\%  & 55.56\%     & 88.89\%  & 77.78\%      \\
Isomer Decoys    & 88.89\%  & 66.67\%     & 100.00\% & 100.00\%     \\
\hline
                 & CFM-ID   & MetFrag     & CMSSP    & CSU-MS2      \\
\hline
MetaCyc Database & 88.89\%  & 88.89\%     & 55.56\%  & 88.89\%      \\
Isomer Decoys    & 100.00\% & 88.89\%     & 77.78\%  & 100.00\%     \\
\hline
\end{tabular}
\end{table}

\begin{table}[]
\caption{Identification accuracies of discriminators during training for database searching against MetaCyc (GAN-0 is trained using structural isomers from PubChem as negative samples, and GAN-9 corresponds to MS2MetGAN)}
\label{tab:MetaCyc-GANs}
\begin{tabular}{lccccc}
\hline
            & GAN-0    & GAN-1   & GAN-2   & GAN-3   & GAN-4   \\
\hline
CASMI2016SP & 80.33\%  & 49.18\% & 65.57\% & 61.48\% & 67.21\% \\
CASMI2017SP & 52.38\%  & 33.33\% & 61.90\% & 61.90\% & 57.14\% \\
CASMI2016FP & 91.08\%  & 43.04\% & 72.97\% & 65.35\% & 72.44\% \\
CASMI2017FP & 100.00\% & 52.74\% & 72.60\% & 69.18\% & 67.12\% \\
CASMI2022P  & 33.54\%  & 47.83\% & 33.54\% & 35.40\% & 34.78\% \\
GNPS-S      & 98.96\%  & 43.01\% & 67.36\% & 55.44\% & 62.18\% \\
GNPS-M      & 96.91\%  & 62.37\% & 74.23\% & 34.54\% & 69.59\% \\
EMBL-MCF    & 7.07\%   & 4.82\%  & 35.37\% & 35.37\% & 64.63\% \\
MoNA        & 64.91\%  & 39.47\% & 48.25\% & 18.42\% & 57.89\% \\
Mean        & 69.46\%  & 41.75\% & 59.09\% & 48.56\% & 61.44\% \\
SD\footnotemark[1]          & 0.3286   & 0.1613  & 0.1603  & 0.1786  & 0.1121  \\
\hline
            & GAN-5    & GAN-6   & GAN-7   & GAN-8   & GAN-9   \\
\hline
CASMI2016SP & 74.59\%  & 74.59\% & 81.15\% & 74.59\% & 81.97\% \\
CASMI2017SP & 83.33\%  & 69.05\% & 88.10\% & 80.95\% & 88.10\% \\
CASMI2016FP & 80.84\%  & 82.15\% & 85.83\% & 81.89\% & 85.56\% \\
CASMI2017FP & 73.29\%  & 76.03\% & 81.51\% & 80.14\% & 82.19\% \\
CASMI2022P  & 42.24\%  & 34.78\% & 37.27\% & 36.02\% & 37.89\% \\
GNPS-S      & 61.66\%  & 63.73\% & 72.02\% & 65.80\% & 73.06\% \\
GNPS-M      & 65.98\%  & 75.26\% & 74.74\% & 75.26\% & 72.16\% \\
EMBL-MCF    & 86.17\%  & 89.39\% & 89.39\% & 88.42\% & 93.25\% \\
MoNA        & 68.42\%  & 71.05\% & 74.56\% & 66.67\% & 72.81\% \\
Mean        & 70.72\%  & 70.67\% & 76.06\% & 72.19\% & 76.33\% \\
SD\footnotemark[1]          & 0.1343   & 0.1536  & 0.1581  & 0.1538  & 0.1618 \\
\hline
\end{tabular}
\footnotetext[1]{SD: Standard Deviation}
\end{table}

\begin{table}[]
\caption{Identification accuracies of discriminators during training for database searching against isomer decoys (GAN-0 is trained using structural isomers from PubChem as negative samples, and GAN-9 corresponds to MS2MetGAN)}
\label{tab:PubChem-GANs}
\begin{tabular}{lccccc}
\hline
            & GAN-0    & GAN-1   & GAN-2   & GAN-3   & GAN-4   \\
\hline
CASMI2016SP & 89.34\%  & 55.74\% & 57.38\% & 68.85\% & 77.05\% \\
CASMI2017SP & 76.19\%  & 16.67\% & 19.05\% & 64.29\% & 59.52\% \\
CASMI2016FP & 92.91\%  & 60.10\% & 62.99\% & 67.19\% & 78.74\% \\
CASMI2017FP & 100.00\% & 45.21\% & 45.89\% & 72.60\% & 76.71\% \\
CASMI2022P  & 19.88\%  & 34.16\% & 22.36\% & 27.33\% & 32.30\% \\
GNPS-S      & 82.90\%  & 37.82\% & 39.90\% & 52.85\% & 63.73\% \\
GNPS-M      & 91.75\%  & 65.46\% & 64.43\% & 73.71\% & 82.47\% \\
EMBL-MCF    & 17.04\%  & 7.40\%  & 9.00\%  & 17.36\% & 47.27\% \\
MoNA        & 66.67\%  & 45.61\% & 42.98\% & 50.88\% & 64.04\% \\
Mean        & 70.74\%  & 40.91\% & 40.44\% & 55.01\% & 64.65\% \\
SD\footnotemark[1]          & 0.3122   & 0.1936  & 0.1994  & 0.2028  & 0.1655  \\
\hline
            & GAN-5    & GAN-6   & GAN-7   & GAN-8   & GAN-9   \\
\hline
CASMI2016SP & 78.69\%  & 93.44\% & 86.07\% & 84.43\% & 86.07\% \\
CASMI2017SP & 83.33\%  & 95.24\% & 92.86\% & 85.71\% & 90.48\% \\
CASMI2016FP & 81.89\%  & 82.15\% & 88.19\% & 87.40\% & 87.40\% \\
CASMI2017FP & 78.08\%  & 92.47\% & 85.62\% & 88.36\% & 86.30\% \\
CASMI2022P  & 31.06\%  & 36.02\% & 39.13\% & 37.27\% & 37.89\% \\
GNPS-S      & 61.14\%  & 74.09\% & 73.58\% & 77.20\% & 75.65\% \\
GNPS-M      & 75.77\%  & 89.18\% & 83.51\% & 85.57\% & 79.90\% \\
EMBL-MCF    & 58.52\%  & 55.31\% & 86.50\% & 90.03\% & 93.25\% \\
MoNA        & 71.05\%  & 71.05\% & 78.07\% & 75.44\% & 77.19\% \\
Mean        & 68.84\%  & 76.55\% & 79.28\% & 79.05\% & 79.35\% \\
SD\footnotemark[1]          & 0.1662   & 0.2003  & 0.1607  & 0.1641  & 0.1663 \\
\hline
\end{tabular}
\footnotetext[1]{SD: Standard Deviation}
\end{table}

\section{Discussion}\label{sec3}

%\subsubsection{This is an example for third level head---subsubsection head}\label{subsubsec2}

We present a new method to improve metabolite identification and, based on this method, develop MS2MetGAN, a database-search tool for identifying metabolites from MS/MS spectra. The method represents compound structures and MS/MS spectra as latent numerical vectors learned by autoencoders. It then employs a GAN in which the generator produces latent structure vectors conditioned on spectrum latent vectors. The discriminator learns to distinguish true latent metabolite–spectrum matches (MSMs), which are pairs of latent vectors from experimentally verified metabolites and their corresponding experimental spectra, from decoy latent MSMs formed by pairing latent vectors of synthetic metabolites with the same experimental spectra. The trained discriminator is then used to rank candidate metabolites from a user-specified database and identify the most likely true metabolite.

This method comprises two key steps: (i) encoding compound structures and MS/MS spectra as latent numerical vectors using autoencoders, and (ii) improving metabolite–spectrum match (MSM) identification using a GAN. Representing spectra and structures as vectors simplifies decoy generation, because the GAN generator can produce latent vectors corresponding to synthetic metabolites directly in the learned latent space. Moreover, the discriminator trained with structural isomers from PubChem as negative samples (GAN-0) achieves higher average performance than all baseline tools. In pairwise comparisons, GAN-0 outperforms all baselines except SF-Matching. This strong performance suggests that the learned latent representations preserve key information from both compound structures and MS/MS spectra. 

A major limitation of MS2MetGAN is the computational cost of pre-encoding experimental spectra and metabolite candidates with the autoencoders. All user-provided MS/MS spectra must be processed by the spectrum autoencoder in decoder-frozen mode (see the ``Autoencoder architectures and training'' section for more details). Similarly, each metabolite candidate must be processed by the structure autoencoder in decoder-frozen mode. These requirements impose additional computational overhead on users. One practical solution is to pre-encode all compounds in a user-specified database when the database will be reused across multiple analyses. In this study, we pre-encoded all compounds in MetaCyc (v29.0) and release their latent vectors. Another solution is to leverage parallel computing for large-scale pre-encoding of spectra and compounds, since each spectrum or compound can be processed independently.

MS2MetGAN achieves the best overall performance compared with the baseline tools. Moreover, its identification accuracy is more balanced across benchmarks. For example, although GAN-0 attains 100.00\% accuracy on CASMI2017FP with both isomeric and MetaCyc decoys, its accuracy on EMBL-MCF is only 17.04\% (isomeric decoys) and 7.07\% (MetaCyc decoys). After nine rounds of GAN-based refinement, MS2MetGAN increases the EMBL-MCF accuracy to 93.25\% under both decoy settings, while reducing performance on CASMI2017FP to 86.30\% (isomeric decoys) and 82.19\% (MetaCyc decoys). The main exception is CASMI2022P, where MS2MetGAN achieves 37.89\% accuracy with both isomeric and MetaCyc decoys. This result may indicate that metabolites in CASMI2022P exhibit fragmentation patterns that differ from those represented in the training dataset. Notably, MIDAS outperforms the other baseline tools on CASMI2022P under the isomeric-decoy setting and achieves the highest average identification accuracy among baseline methods under the MetaCyc-decoy setting. As the only method in our comparison that does not use machine learning, MIDAS assumes uniform fragmentation behavior across chemical bonds, suggesting that this simplifying assumption can remain effective in challenging identification scenarios. Together, these findings support the rationale for exploring integration of MIDAS to complement machine-learning-based metabolite identification approaches in future work.

\section{Methods}\label{sec4}
%\subsection{Problem formulation}
%Database-searching-based Metabolite identification can be considered as a supervised classification problem with data {(x)}. 

\subsection{Autoencoder architectures and training}\label{subsec_autoencoder}
This study trains two autoencoders separately, one for compound structures and one for MS/MS spectra. Each autoencoder consists of an encoder and a decoder. The compound-structure autoencoder adopts the MassFormer architecture \cite{Young2024}. The encoder extracts node, edge, and spatial features and feeds them into a transformer-based network. Its output is mapped to a 1,280-dimensional latent vector via a normalization layer followed by a $10 \times 10$ fully connected layer. The latent vector is then passed to a decoder comprised of a 12-layer transformer, a normalization layer, and a $10 \times 3$ linear layer to reconstruct the node features. The MS/MS spectrum autoencoder is designed to preserve key spectral characteristics through linear projection. Each binned spectrum is a 15,000-dimensional vector (see the ``Training dataset'' section in Supplementary Information for more details). We first reduce it to 7,500 dimensions by summing adjacent elements, and then pass the resulting vector to an encoder with three shared fully connected layers to obtain a 1,500-dimensional latent vector. The use of multiple fully connected layers facilitates the application of dropout during training. The decoder comprises seven linear layers to reconstruct an approximation of the 7,500-dimensional spectrum vector. Hyperbolic tangent (Tanh) activations are used in the first six layers, and a Rectified Linear Unit (ReLU) activation is used in the final layer.

We train both autoencoders using AdamW \cite{Loshchilov2019} with a two-stage strategy. In the first stage, we train the full autoencoder (encoder and decoder) on the entire training dataset until performance no longer improves. The resulting structure and spectrum autoencoders are denoted CA1 and MA1, respectively. In the second stage, we further optimize CA1 and MA1 on a per-sample basis with the decoder parameters fixed. Specifically, each training sample is used to fine-tune its own encoder while sharing a common decoder (the decoder learned in stage 1). During preprocessing of the test data, each test compound candidate or MS/MS spectrum similarly fine-tunes its own encoder initialized from CA1 or MA1, respectively, with the decoder frozen. Because the decoders remain fixed in stage 2, the resulting latent codes are comparable and can be used jointly. Finally, all layers in the spectrum autoencoder apply dropout with probability $1 \times 10^{-4}$ during training.

\subsection{GAN architecture and training}\label{subsec_gan}

A GAN consists of a generator and a discriminator. In this study, the generator takes a spectrum latent code as input and outputs a latent code representing a compound structure. The generator comprises three linear layers with ReLU activations. The discriminator takes as input the concatenation of a paired spectrum latent code and compound latent code. For a true metabolite–spectrum match (MSM), the target output is 1. For a decoy MSM, the target output is 0. The discriminator comprises a 16-layer transformer followed by a normalization layer, a shared fully connected layer, and a final linear layer with a ReLU activation.  

Both the generator and discriminator are trained using AdamW \cite{Loshchilov2019}. In the initial round (round 0), we train only the discriminator to distinguish true MSMs from decoy MSMs, where decoy compound latent codes are derived from PubChem isomers (see the ``Training dataset'' section in Supplementary Information for more details). Training stops when discriminator accuracy reaches 99\%. The resulting discriminator is denoted GAN-0. Starting from round 1, we alternately train the generator and discriminator. The generator is trained to produce a synthetic compound latent vector conditioned on each input spectrum latent vector. 
For all rounds except the final round, generator training stops when 99\% of generated input–output pairs fool the discriminator from the previous round. In the final round, the generator achieves a fooling rate of 98.75\%. In the same round, the discriminator is trained to distinguish true MSMs from decoy MSMs, where decoy compound latent codes are generated by the current-round generator. Discriminator training stops when accuracy reaches 99\%. We repeat this procedure through round 9. The discriminator trained in round $k (k=1,2,...,9)$ is denoted GAN-$k$, and GAN-9 corresponds to MS2MetGAN.

\section{Conclusion}\label{sec13}

In this study, we present a new database-search approach for identifying metabolites from MS/MS spectra. Our method encodes metabolite structures and MS/MS spectra into latent vectors using separate autoencoders, enabling scoring to be performed directly in the latent domain. We then use adversarial training to generate informative decoy metabolite latents and to strengthen the discriminator used for metabolite–spectrum matching. Building on this framework, we develop MS2MetGAN, our database-search tool for metabolite identification. By improving the quality of negative examples and refining the matching score, MS2MetGAN improves overall ranking performance and outperforms existing metabolite identification methods on average across all benchmarks.

\backmatter

\bmhead{Supplementary information}

Supplementary Information is provided as a separate PDF.

\bmhead{Acknowledgements}
This work was performed with partial support from the National Science Foundation under Grant No. 2053286 and from the National Artificial Intelligence Research Resource (NAIRR) Pilot under Grant No. NAIRR240213.

\bmhead{Web Server}
http://bioinfo3.research.utc.edu/MS2MetGan/

%Acknowledgements are not compulsory. Where included they should be brief. Grant or contribution numbers may be acknowledged.

%Please refer to Journal-level guidance for any specific requirements.

%\section*{Declarations}

%Some journals require declarations to be submitted in a standardised format. Please check the Instructions for Authors of the journal to which you are submitting to see if you need to complete this section. If yes, your manuscript must contain the following sections under the heading `Declarations':

%\begin{itemize}
%\item Conflict of interest/Competing interests (check journal-specific guidelines for which heading to use)
%\item Ethics approval and consent to participate
%\item Consent for publication
%\item Data availability 
%\end{itemize}

\noindent
%If any of the sections are not relevant to your manuscript, please include the heading and write `Not applicable' for that section. 



\bibliography{sn-bibliography}% common bib file

\end{document}

\title{Supplementary Information for: \\
MS2MetGAN: Latent-space adversarial training for metabolite–spectrum matching in MS/MS database search}
\author{Meng Tsai, Alexzander Dwyer, Estelle Nuckels, Yingfeng Wang*}
\maketitle

\section{Supplementary Text}
% Additional methods, ablation studies, extra experiments, etc.

\subsection{Datasets and data preprocessing}\label{subsec_data}
\subsubsection{Training dataset}\label{subsubsec_training_data}
To train MS2MetGAN, we extract 51,934 [M+H]\textsuperscript{+} MS/MS spectra and their associated metabolite information from the MassBank of North America (MoNA). Spectra corresponding to the same metabolite are combined by concatenating their peak lists (i.e., appending peaks without merging or summing intensities). After this consolidation, the dataset contains 10,566 spectra. To convert spectra into bin-based embeddings, we partition each spectrum into m/z bins of width 0.1 over the range \(m/z \leq 1500\). Peaks with \(m/z > 1500\) are ignored. For each bin, the intensity is set to 0 if no peak falls within the bin. Otherwise, the intensities of all peaks in that bin are summed to obtain the bin intensity. The resulting bin-intensity vector is then normalized to the range \([0, 1]\). To construct negative samples for initial discriminator training, we retrieved one structural isomer decoy per spectrum from PubChem. For each spectrum’s associated metabolite, PubChem was queried for a compound with the same chemical formula and a MACCS key fingerprint similarity no more than 0.75. Each decoy was paired with the corresponding true standard spectrum for initial training of the discriminator.

%MACCS keys fingerprint similarity

\subsubsection{Test datasets}\label{subsubsec_test_data}
This study uses widely adopted public test datasets of MS/MS spectra acquired in positive ion mode ([M+H]\textsuperscript{+}) and their associated compounds to evaluate existing metabolite identification tools and our approach. CASMI contests provide public standard MS/MS spectra with associated compound annotations. We use the positive-ion datasets from CASMI 2016, CASMI 2017, and CASMI 2022, referred to as CASMI2016FP, CASMI2017FP, and CASMI2022P, respectively. Notably, we remove 16 spectra from CASMI2022 because their associated compounds appear in the training dataset. SF-Matching \cite{Li2020} evaluates on subsets of CASMI2016FP and CASMI2017FP, denoted CASMI2016SP and CASMI2017SP. Therefore, we also include these two benchmark sets. SF-Matching \cite{Li2020} and MolDiscovery \cite{Cao2021} additionally evaluate on positive-ion datasets from GNPS \cite{Wang2016}, referred to as GNPS-S and GNPS-M, respectively. We include both in our evaluation. Finally, we use the EMBL-MCF and MoNA benchmarks, which are also employed by SF-Matching \cite{Li2020} and MolDiscovery \cite{Cao2021}, respectively. There is no spectra and associated compounds overlap between the training and test datasets. This study does not merge test MS/MS spectra. All test spectra are binned using the same procedure as for the training spectra. 

As for each test benchmark, this study uses two compound databases. The first database includes all compounds in MetaCyc (v29.0) together with the true compounds in the benchmark set. The second database contains the true compounds and their isomer decoys. For each spectrum, we query PubChem for isomeric compounds with the same chemical formula as the corresponding true compound, filter out candidates with a MACCS key fingerprint similarity greater than 0.75, and then retain up to ten remaining isomers per true compound. This similarity threshold is consistent with the criterion used to construct decoys in the training dataset.

%MassBank of North America (MoNA) \cite{mona_rrid}.

%GNPS \cite{Wang2016}

%MetaCyc \cite{Caspi2014}

%PubChem \cite{Kim2025}

\subsubsection{Baseline tools}\label{subsubsec_baseline_tools}
Our evaluation focuses on tools that use a database-search strategy. Most recent database-search–based methods incorporate machine-learning models to improve candidate ranking. MAGMa \cite{Ridder2012} scores metabolite–spectrum matches using a fragmentation-based scoring scheme with predefined penalty parameters. MAGMa+ \cite{Verdegem2016} optimizes selected penalty parameters and employs random forest classifiers to further improve identification performance. MetFrag \cite{Ruttkies2019} scores metabolite–spectrum matches by enumerating plausible fragments of each candidate compound and matching them to peaks in the spectrum. It then refines these scores using probabilities of fragment and neutral-loss fingerprints conditioned on mass-to-charge ratios, learned from reference metabolite–spectrum pairs. CSI:FingerID \cite{Dührkop2015} uses a set of support vector machines to predict a molecular fingerprint from an MS/MS spectrum and its inferred fragmentation tree. Candidate metabolites are then scored by comparing their structural fingerprints to the predicted fingerprint, and this similarity defines the metabolite–spectrum match score. CSI:FingerID is implemented and deployed within SIRIUS \cite{Dührkop2019}. SF-Matching \cite{Li2020} trains random forest models that, from the chemical fingerprint of a candidate metabolite, estimate the probabilities of observing specific fragment peaks. Metabolite–spectrum matches are then scored by summing the experimental peak intensities weighted by these predicted peak probabilities. CFM-ID \cite{Wang2021} enumerates plausible fragments of a given metabolite and encodes each fragment using atom-level, bond-level, and topological features. These feature representations are fed into a neural network that learns a mapping from molecular fragment characteristics to the likelihood that each fragment will appear in an MS/MS spectrum. The trained model is then used to score metabolite–spectrum matches based on the agreement between predicted and observed fragment ions. MolDiscovery \cite{Cao2021} learns a probabilistic fragmentation model from known small-molecule–spectrum pairs and then applies this model to compute likelihood-based scores for metabolite–spectrum matches. However, MolDiscovery is currently not available with public source code or batch-processing support needed for large-scale evaluation. Therefore, this study excludes MolDiscovery from the performance comparison. CMSSP \cite{Chen2024} encodes molecular structures and MS/MS spectra as numerical vectors using a graph neural network–based structure encoder (augmented with molecular fingerprints) and a Transformer-based spectrum encoder, respectively. The resulting embeddings are passed through small projection networks, and dot products between the projected spectrum and structure embeddings form a similarity matrix on which a contrastive loss is applied during training. At inference, these similarities are then used to score metabolite–spectrum pairs. More recent representation-learning methods embed molecular structures and spectra into a shared space. CSU-MS2 \cite{Xie2025} employs dedicated spectral and molecular encoders to represent MS/MS spectra and molecular structures, respectively, with both encoders incorporating External Space Attention Aggregation modules to better align the two modalities. The model uses contrastive learning to train numerical embeddings for spectra and candidate metabolites. At inference time, metabolite–spectrum matches are scored by computing the cosine similarity between their embeddings, and the final score is obtained by aggregating similarities from models trained for different collision-energy settings. One notable exception is MIDAS \cite{Wang2014}, whose scoring model does not use machine learning. MIDAS constructs a three-level fragmentation tree by systematically disconnecting each linear bond and cleaving open each ring (excluding bonds involving hydrogen). For each fragment, all possible fragment–peak matches are scored, and the metabolite–spectrum match score is obtained by summing, over all peaks, the highest-scoring fragment–peak match.

%\clearpage
%\section*{Supplementary Figures}

%\begin{figure}[h]
%  \centering
%  \includegraphics[width=0.9\linewidth]{figS1.pdf}
%  \caption{Caption for Supplementary Fig.~S1.}
%  \label{fig:S1}
%\end{figure}

\clearpage
\section{Supplementary Tables}

\begin{table}[h]
%\caption{MetaCyc MIDAS}
\caption{Distribution of the true-compound rank in MIDAS database searches against MetaCyc across benchmark spectra datasets}
\label{tab:MetaCyc-MIDAS}
\begin{tabular}{lccccl}
\hline
                       & CASMI2016SP & CASMI2017SP & CASMI2016FP & CASMI2017FP & \multicolumn{1}{c}{CASMI2022P} \\
\hline
total                  & 122         & 42          & 381         & 146         & \multicolumn{1}{c}{161}        \\
rank 1                 & 111         & 37          & 324         & 98          & \multicolumn{1}{c}{111}        \\
rank 2                 & 5           & 1           & 27          & 7           & \multicolumn{1}{c}{24}         \\
rank 3                 & 3           & 0           & 12          & 9           & \multicolumn{1}{c}{9}          \\
rank 4                 & 0           & 1           & 3           & 9           & \multicolumn{1}{c}{6}          \\
rank 5                 & 0           & 1           & 3           & 6           & \multicolumn{1}{c}{1}          \\
ranks $\ge 6$ & 3           & 2           & 12          & 17          & \multicolumn{1}{c}{10}         \\
\hline
\end{tabular}
\end{table}

\begin{table}[h]
%\caption{MetaCyc SF-Matching}
\caption{Distribution of the true-compound rank in SF-Matching database searches against MetaCyc across benchmark spectra datasets}
\label{tab:MetaCyc-SF-Matching}
\begin{tabular}{lccccl}
\hline
                       & CASMI2016SP & CASMI2017SP & CASMI2016FP & CASMI2017FP & \multicolumn{1}{c}{CASMI2022P} \\
\hline
total                  & 122         & 42          & 381         & 146         & \multicolumn{1}{c}{161}        \\
rank 1                 & 121         & 35          & 374         & 90          & \multicolumn{1}{c}{94}         \\
rank 2                 & 0           & 4           & 1           & 20          & \multicolumn{1}{c}{30}         \\
rank 3                 & 0           & 0           & 3           & 8           & \multicolumn{1}{c}{18}         \\
rank 4                 & 0           & 1           & 0           & 5           & \multicolumn{1}{c}{5}          \\
rank 5                 & 0           & 2           & 1           & 6           & \multicolumn{1}{c}{4}          \\
ranks $\ge 6$ & 1           & 0           & 2           & 17          & \multicolumn{1}{c}{10}         \\
\hline
\end{tabular}
\end{table}

\begin{table}[h]
%\caption{MetaCyc MAGMA+}
\caption{Distribution of the true-compound rank in MAGMA+ database searches against MetaCyc across benchmark spectra datasets}
\label{tab:MetaCyc-MAGMA}
\begin{tabular}{lccccc}
\hline
                       & CASMI2016SP & CASMI2017SP & CASMI2016FP & CASMI2017FP & CASMI2022P \\
\hline
total                  & 122         & 42          & 381         & 146         & 161        \\
rank 1                 & 67          & 25          & 218         & 75          & 87         \\
rank 2                 & 10          & 5           & 40          & 17          & 25         \\
rank 3                 & 2           & 2           & 12          & 7           & 14         \\
rank 4                 & 1           & 0           & 2           & 4           & 8          \\
rank 5                 & 1           & 0           & 13          & 3           & 3          \\
ranks $\ge 6$ & 41          & 10          & 96          & 40          & 24         \\
\hline
\end{tabular}
\end{table}

\begin{table}[h]
%\caption{MetaCyc CSI:FingerID}
\caption{Distribution of the true-compound rank in CSI:FingerID database searches against MetaCyc across benchmark spectra datasets}
\label{tab:MetaCyc-CSIFingerID}
\begin{tabular}{lccccl}
\hline
                       & CASMI2016SP & CASMI2017SP & CASMI2016FP & CASMI2017FP & \multicolumn{1}{c}{CASMI2022P} \\
\hline
total                  & 122         & 42          & 381         & 146         & \multicolumn{1}{c}{161}        \\
rank 1                 & 73          & 27          & 331         & 45          & \multicolumn{1}{c}{125}        \\
rank 2                 & 0           & 2           & 32          & 14          & \multicolumn{1}{c}{11}         \\
rank 3                 & 0           & 2           & 3           & 17          & \multicolumn{1}{c}{4}          \\
rank 4                 & 0           & 0           & 3           & 4           & \multicolumn{1}{c}{3}          \\
rank 5                 & 0           & 1           & 1           & 5           & \multicolumn{1}{c}{1}          \\
ranks $\ge 6$ & 49          & 10          & 11          & 61          & \multicolumn{1}{c}{17}         \\
\hline

\end{tabular}
\end{table}

\begin{table}[h]
%\caption{MetaCyc CFM-ID}
\caption{Distribution of the true-compound rank in CFM-ID database searches against MetaCyc across benchmark spectra datasets}
\label{tab:MetaCyc-CFM-ID}
\begin{tabular}{lccccl}
\hline
                       & CASMI2016SP & CASMI2017SP & CASMI2016FP & CASMI2017FP & \multicolumn{1}{c}{CASMI2022P} \\
\hline
total                  & 122         & 42          & 381         & 146         & \multicolumn{1}{c}{161}        \\
rank 1                 & 85          & 26          & 253         & 62          & \multicolumn{1}{c}{106}        \\
rank 2                 & 14          & 5           & 53          & 28          & \multicolumn{1}{c}{28}         \\
rank 3                 & 6           & 3           & 27          & 12          & \multicolumn{1}{c}{6}          \\
rank 4                 & 6           & 2           & 17          & 10          & \multicolumn{1}{c}{6}          \\
rank 5                 & 4           & 1           & 8           & 9           & \multicolumn{1}{c}{5}          \\
ranks $\ge 6$ & 7           & 5           & 23          & 25          & \multicolumn{1}{c}{10}         \\
\hline
\end{tabular}
\end{table}

\begin{table}[h]
%\caption{MetaCyc MetFrag}
\caption{Distribution of the true-compound rank in MetFrag database searches against MetaCyc across benchmark spectra datasets}
\label{tab:MetaCyc-MetFrag}
\begin{tabular}{lccccl}
\hline
                       & CASMI2016SP & CASMI2017SP & CASMI2016FP & CASMI2017FP & \multicolumn{1}{c}{CASMI2022P} \\
\hline
total                  & 122         & 42          & 381         & 146         & \multicolumn{1}{c}{161}        \\
rank 1                 & 52          & 11          & 129         & 26          & \multicolumn{1}{c}{105}        \\
rank 2                 & 14          & 2           & 44          & 7           & \multicolumn{1}{c}{25}         \\
rank 3                 & 4           & 5           & 18          & 10          & \multicolumn{1}{c}{12}         \\
rank 4                 & 4           & 4           & 12          & 8           & \multicolumn{1}{c}{6}          \\
rank 5                 & 5           & 2           & 11          & 4           & \multicolumn{1}{c}{2}          \\
ranks $\ge 6$ & 43          & 18          & 167         & 91          & \multicolumn{1}{c}{11}         \\
\hline          
\end{tabular}
\end{table}

\begin{table}[h]
%\caption{MetaCyc CMSSP}
\caption{Distribution of the true-compound rank in CMSSP database searches against MetaCyc across benchmark spectra datasets}
\label{tab:MetaCyc-CMSSP}
\begin{tabular}{lccccl}
\hline
                       & CASMI2016SP & CASMI2017SP & CASMI2016FP & CASMI2017FP & \multicolumn{1}{c}{CASMI2022P} \\
\hline
total                  & 122         & 42          & 381         & 146         & \multicolumn{1}{c}{161}        \\
rank 1                 & 108         & 39          & 324         & 110         & \multicolumn{1}{c}{72}         \\
rank 2                 & 10          & 0           & 29          & 10          & \multicolumn{1}{c}{25}         \\
rank 3                 & 0           & 1           & 5           & 6           & \multicolumn{1}{c}{8}          \\
rank 4                 & 0           & 1           & 7           & 0           & \multicolumn{1}{c}{9}          \\
rank 5                 & 1           & 0           & 5           & 1           & \multicolumn{1}{c}{4}          \\
ranks $\ge 6$ & 3           & 1           & 11          & 19          & \multicolumn{1}{c}{43}         \\
\hline
\end{tabular}
\end{table}

\begin{table}[h]
%\caption{MetaCyc CSU-MS2}
\caption{Distribution of the true-compound rank in CSU-MS2 database searches against MetaCyc across benchmark spectra datasets}
\label{tab:MetaCyc-CSU-MS2}
\begin{tabular}{lccccl}
\hline
                       & CASMI2016SP & CASMI2017SP & CASMI2016FP & CASMI2017FP & \multicolumn{1}{c}{CASMI2022P} \\
\hline
total                  & 122         & 42          & 381         & 146         & \multicolumn{1}{c}{161}        \\
rank 1                 & 32          & 23          & 93          & 44          & \multicolumn{1}{c}{67}         \\
rank 2                 & 20          & 7           & 56          & 31          & \multicolumn{1}{c}{27}         \\
rank 3                 & 10          & 4           & 43          & 12          & \multicolumn{1}{c}{15}         \\
rank 4                 & 11          & 3           & 32          & 10          & \multicolumn{1}{c}{11}         \\
rank 5                 & 4           & 0           & 27          & 15          & \multicolumn{1}{c}{11}         \\
ranks $\ge 6$ & 45          & 5           & 130         & 34          & \multicolumn{1}{c}{30}         \\
\hline          
\end{tabular}
\end{table}

\begin{table}[h]
%\caption{MetaCyc MS2MetGAN}
\caption{Distribution of the true-compound rank in MS2MetGAN database searches against MetaCyc across benchmark spectra datasets}
\label{tab:MetaCyc-MS2MetGAN}
\begin{tabular}{lccccc}
\hline
                       & CASMI2016SP & CASMI2017SP & CASMI2016FP & CASMI2017FP & CASMI2022P \\
\hline
total                  & 122         & 42          & 381         & 146         & 161        \\
rank 1                 & 100         & 37          & 326         & 120         & 61         \\
rank 2                 & 0           & 0           & 0           & 1           & 21         \\
rank 3                 & 0           & 0           & 0           & 0           & 24         \\
rank 4                 & 0           & 0           & 0           & 0           & 8          \\
rank 5                 & 0           & 0           & 0           & 0           & 13         \\
ranks $\ge 6$ & 22          & 5           & 55          & 25          & 34         \\
\hline
\end{tabular}
\end{table}

\begin{table}[h]
%\caption{PubChem MIDAS}
\caption{Distribution of the true-compound rank in MIDAS database searches against isomer decoys across benchmark spectra datasets}
\label{tab:PubChem-MIDAS}
\begin{tabular}{lccccc}
\hline
                       & CASMI2016SP & CASMI2017SP & CASMI2016FP & CASMI2017FP & CASMI2022P \\
\hline
total                  & 122         & 42          & 381         & 146         & 161        \\
rank 1                 & 67          & 24          & 208         & 58          & 69         \\
rank 2                 & 16          & 4           & 39          & 17          & 12         \\
rank 3                 & 8           & 5           & 28          & 12          & 8          \\
rank 4                 & 3           & 1           & 13          & 10          & 10         \\
rank 5                 & 6           & 0           & 23          & 6           & 9          \\
ranks $\ge 6$ & 22          & 8           & 70          & 43          & 53         \\
\hline
\end{tabular}
\end{table}

\begin{table}[h]
%\caption{PubChem SF-Matching}
\caption{Distribution of the true-compound rank in SF-Matching database searches against isomer decoys across benchmark spectra datasets}
\label{tab:PubChem-SF-Matching}
\begin{tabular}{lccccl}
\hline
                       & CASMI2016SP & CASMI2017SP & CASMI2016FP & CASMI2017FP & \multicolumn{1}{c}{CASMI2022P} \\
\hline
total                  & 122         & 42          & 381         & 146         & \multicolumn{1}{c}{161}        \\
rank 1                 & 120         & 34          & 363         & 88          & \multicolumn{1}{c}{55}         \\
rank 2                 & 0           & 4           & 9           & 12          & \multicolumn{1}{c}{15}         \\
rank 3                 & 0           & 1           & 3           & 7           & \multicolumn{1}{c}{12}         \\
rank 4                 & 1           & 1           & 2           & 4           & \multicolumn{1}{c}{6}          \\
rank 5                 & 0           & 1           & 1           & 4           & \multicolumn{1}{c}{6}          \\
ranks $\ge 6$ & 1           & 1           & 3           & 31          & \multicolumn{1}{c}{67}         \\
\hline
\end{tabular}
\end{table}

\begin{table}[h]
%\caption{PubChem MAGMA+}
\caption{Distribution of the true-compound rank in MAGMA+ database searches against isomer decoys across benchmark spectra datasets}
\label{tab:PubChem-MAGMA}
\begin{tabular}{lccccc}
\hline
                       & CASMI2016SP & CASMI2017SP & CASMI2016FP & CASMI2017FP & CASMI2022P \\
\hline
total                  & 122         & 42          & 381         & 146         & 161        \\
rank 1                 & 40          & 13          & 126         & 40          & 48         \\
rank 2                 & 14          & 4           & 46          & 12          & 12         \\
rank 3                 & 6           & 5           & 25          & 15          & 10         \\
rank 4                 & 6           & 0           & 18          & 7           & 9          \\
rank 5                 & 1           & 1           & 18          & 6           & 7          \\
ranks $\ge 6$ & 55          & 19          & 148         & 66          & 75         \\
\hline
\end{tabular}
\end{table}

\begin{table}[h]
%\caption{PubChem CSI:FingerID}
\caption{Distribution of the true-compound rank in CSI:FingerID database searches against isomer decoys across benchmark spectra datasets}
\label{tab:PubChem-CSIFingerID}
\begin{tabular}{lccccl}
\hline
                       & CASMI2016SP & CASMI2017SP & CASMI2016FP & CASMI2017FP & \multicolumn{1}{c}{CASMI2022P} \\
\hline
total                  & 122         & 42          & 381         & 146         & \multicolumn{1}{c}{161}        \\
rank 1                 & 71          & 29          & 320         & 46          & \multicolumn{1}{c}{47}         \\
rank 2                 & 1           & 0           & 13          & 4           & \multicolumn{1}{c}{12}         \\
rank 3                 & 0           & 0           & 3           & 1           & \multicolumn{1}{c}{5}          \\
rank 4                 & 0           & 0           & 3           & 1           & \multicolumn{1}{c}{5}          \\
rank 5                 & 0           & 0           & 0           & 2           & \multicolumn{1}{c}{4}          \\
ranks $\ge 6$ & 50          & 13          & 42          & 92          & \multicolumn{1}{c}{88}         \\
\hline          
\end{tabular}
\end{table}

\begin{table}[h]
%\caption{PubChem CFM-ID}
\caption{Distribution of the true-compound rank in CFM-ID database searches against isomer decoys across benchmark spectra datasets}
\label{tab:PubChem-CFM-ID}
\begin{tabular}{lccccl}
\hline
                       & CASMI2016SP & CASMI2017SP & CASMI2016FP & CASMI2017FP & \multicolumn{1}{c}{CASMI2022P} \\
\hline
total                  & 122         & 42          & 381         & 146         & \multicolumn{1}{c}{161}        \\
rank 1                 & 75          & 19          & 219         & 62          & \multicolumn{1}{c}{56}         \\
rank 2                 & 10          & 7           & 33          & 20          & \multicolumn{1}{c}{14}         \\
rank 3                 & 8           & 2           & 25          & 6           & \multicolumn{1}{c}{6}          \\
rank 4                 & 4           & 0           & 17          & 3           & \multicolumn{1}{c}{6}          \\
rank 5                 & 7           & 1           & 11          & 6           & \multicolumn{1}{c}{5}          \\
ranks $\ge 6$ & 18          & 13          & 76          & 49          & \multicolumn{1}{c}{74}         \\
\hline          
\end{tabular}
\end{table}

\begin{table}[h]
%\caption{PubChem MetFrag}
\caption{Distribution of the true-compound rank in MetFrag database searches against isomer decoys across benchmark spectra datasets}
\label{tab:PubChem-MetFrag}
\begin{tabular}{lccccl}
\hline
                       & CASMI2016SP & CASMI2017SP & CASMI2016FP & CASMI2017FP & \multicolumn{1}{c}{CASMI2022P} \\
\hline
total                  & 122         & 42          & 381         & 146         & \multicolumn{1}{c}{161}        \\
rank 1                 & 63          & 13          & 194         & 71          & \multicolumn{1}{c}{62}         \\
rank 2                 & 13          & 0           & 49          & 16          & \multicolumn{1}{c}{21}         \\
rank 3                 & 11          & 2           & 32          & 12          & \multicolumn{1}{c}{13}         \\
rank 4                 & 1           & 2           & 16          & 10          & \multicolumn{1}{c}{11}         \\
rank 5                 & 2           & 1           & 17          & 7           & \multicolumn{1}{c}{5}          \\
ranks $\ge 6$ & 32          & 24          & 73          & 30          & \multicolumn{1}{c}{49}         \\
\hline          
\end{tabular}
\end{table}

\begin{table}[h]
%\caption{PubChem CMSSP}
\caption{Distribution of the true-compound rank in CMSSP database searches against isomer decoys across benchmark spectra datasets}
\label{tab:PubChem-CMSSP}
\begin{tabular}{lccccl}
\hline
                       & CASMI2016SP & CASMI2017SP & CASMI2016FP & CASMI2017FP & \multicolumn{1}{c}{CASMI2022P} \\
\hline
total                  & 122         & 42          & 381         & 146         & \multicolumn{1}{c}{161}        \\
rank 1                 & 109         & 25          & 342         & 117         & \multicolumn{1}{c}{39}         \\
rank 2                 & 7           & 15          & 18          & 6           & \multicolumn{1}{c}{27}         \\
rank 3                 & 0           & 2           & 7           & 7           & \multicolumn{1}{c}{11}         \\
rank 4                 & 1           & 0           & 2           & 2           & \multicolumn{1}{c}{8}          \\
rank 5                 & 1           & 0           & 2           & 3           & \multicolumn{1}{c}{3}          \\
ranks $\ge 6$ & 4           & 0           & 10          & 11          & \multicolumn{1}{c}{73}         \\
\hline          
\end{tabular}
\end{table}

\begin{table}[h]
%\caption{PubChem CSU-MS2}
\caption{Distribution of the true-compound rank in CSU-MS2 database searches against isomer decoys across benchmark spectra datasets}
\label{tab:PubChem-CSU-MS2}
\begin{tabular}{lccccl}
\hline
                       & CASMI2016SP & CASMI2017SP & CASMI2016FP & CASMI2017FP & \multicolumn{1}{c}{CASMI2022P} \\
\hline
total                  & 122         & 42          & 381         & 146         & \multicolumn{1}{c}{161}        \\
rank 1                 & 9           & 7           & 29          & 30          & \multicolumn{1}{c}{19}         \\
rank 2                 & 13          & 2           & 21          & 7           & \multicolumn{1}{c}{7}          \\
rank 3                 & 7           & 4           & 21          & 7           & \multicolumn{1}{c}{14}         \\
rank 4                 & 7           & 5           & 21          & 12          & \multicolumn{1}{c}{6}          \\
rank 5                 & 3           & 3           & 19          & 8           & \multicolumn{1}{c}{12}         \\
ranks $\ge 6$ & 83          & 21          & 270         & 82          & \multicolumn{1}{c}{103}        \\
\hline          
\end{tabular}
\end{table}

\begin{table}[h]
%\caption{PubChem MS2MetGAN}
\caption{Distribution of the true-compound rank in MS2MetGAN database searches against isomer decoys across benchmark spectra datasets}
\label{tab:PubChem-MS2MetGAN}
\begin{tabular}{lccccl}
\hline
                       & CASMI2016SP & CASMI2017SP & CASMI2016FP & CASMI2017FP & \multicolumn{1}{c}{CASMI2022P} \\
\hline
total                  & 122         & 42          & 381         & 146         & \multicolumn{1}{c}{161}        \\
rank 1                 & 105         & 38          & 333         & 126         & \multicolumn{1}{c}{61}         \\
rank 2                 & 1           & 1           & 1           & 2           & \multicolumn{1}{c}{13}         \\
rank 3                 & 2           & 0           & 2           & 0           & \multicolumn{1}{c}{9}          \\
rank 4                 & 0           & 0           & 1           & 0           & \multicolumn{1}{c}{8}          \\
rank 5                 & 2           & 0           & 5           & 0           & \multicolumn{1}{c}{4}          \\
ranks $\ge 6$ & 12          & 3           & 39          & 18          & \multicolumn{1}{c}{66}         \\
\hline
\end{tabular}
\end{table}

\begin{table}[h]
%\caption{MetaCyc GAN-0}
\caption{Distribution of the true-compound rank in GAN-0 database searches against MetaCyc across benchmark spectral datasets (GAN-0 is trained using structural isomers from PubChem as negative samples)}
\label{tab:MetaCyc-GAN-0}
\begin{tabular}{lccccl}
\hline
                       & CASMI2016SP & CASMI2017SP & CASMI2016FP & CASMI2017FP & \multicolumn{1}{c}{CASMI2022P} \\
\hline
total                  & 122         & 42          & 381         & 146         & \multicolumn{1}{c}{161}        \\
rank 1                 & 98          & 22          & 347         & 146         & \multicolumn{1}{c}{54}         \\
rank 2                 & 0           & 0           & 0           & 0           & \multicolumn{1}{c}{30}         \\
rank 3                 & 0           & 0           & 1           & 0           & \multicolumn{1}{c}{18}         \\
rank 4                 & 0           & 0           & 3           & 0           & \multicolumn{1}{c}{10}         \\
rank 5                 & 1           & 0           & 1           & 0           & \multicolumn{1}{c}{15}         \\
ranks $\ge 6$ & 23          & 20          & 29          & 0           & \multicolumn{1}{c}{34}         \\
\hline          
\end{tabular}
\end{table}

\begin{table}[h]
%\caption{MetaCyc GAN-1}
\caption{Distribution of the true-compound rank in GAN-1 database searches against MetaCyc across benchmark spectral datasets}
\label{tab:MetaCyc-GAN-1}
\begin{tabular}{lccccl}
\hline
                       & CASMI2016SP & CASMI2017SP & CASMI2016FP & CASMI2017FP & \multicolumn{1}{c}{CASMI2022P} \\
\hline
total                  & 122         & 42          & 381         & 146         & \multicolumn{1}{c}{161}        \\
rank 1                 & 60          & 14          & 164         & 77          & \multicolumn{1}{c}{77}         \\
rank 2                 & 2           & 0           & 11          & 2           & \multicolumn{1}{c}{22}         \\
rank 3                 & 0           & 0           & 2           & 1           & \multicolumn{1}{c}{21}         \\
rank 4                 & 2           & 0           & 2           & 0           & \multicolumn{1}{c}{5}          \\
rank 5                 & 0           & 0           & 0           & 1           & \multicolumn{1}{c}{7}          \\
ranks $\ge 6$ & 58          & 28          & 202         & 65          & \multicolumn{1}{c}{29}         \\
\hline
\end{tabular}
\end{table}

\begin{table}[h]
%\caption{MetaCyc GAN-2}
\caption{Distribution of the true-compound rank in GAN-2 database searches against MetaCyc across benchmark spectral datasets}
\label{tab:MetaCyc GAN-2}
\begin{tabular}{lccccl}
\hline
                       & CASMI2016SP & CASMI2017SP & CASMI2016FP & CASMI2017FP & \multicolumn{1}{c}{CASMI2022P} \\
\hline
total                  & 122         & 42          & 381         & 146         & \multicolumn{1}{c}{161}        \\
rank 1                 & 80          & 26          & 278         & 106         & \multicolumn{1}{c}{54}         \\
rank 2                 & 0           & 0           & 3           & 0           & \multicolumn{1}{c}{29}         \\
rank 3                 & 0           & 0           & 1           & 0           & \multicolumn{1}{c}{18}         \\
rank 4                 & 0           & 1           & 1           & 1           & \multicolumn{1}{c}{10}         \\
rank 5                 & 0           & 0           & 1           & 0           & \multicolumn{1}{c}{19}         \\
ranks $\ge 6$ & 42          & 15          & 97          & 39          & \multicolumn{1}{c}{31}         \\
\hline          
\end{tabular}
\end{table}

\begin{table}[h]
%\caption{MetaCyc GAN-3}
\caption{Distribution of the true-compound rank in GAN-3 database searches against MetaCyc across benchmark spectral datasets}
\label{tab:MetaCyc-GAN-3}
\begin{tabular}{lccccl}
\hline
                       & CASMI2016SP & CASMI2017SP & CASMI2016FP & CASMI2017FP & \multicolumn{1}{c}{CASMI2022P} \\
\hline
total                  & 122         & 42          & 381         & 146         & \multicolumn{1}{c}{161}        \\
rank 1                 & 75          & 26          & 249         & 101         & \multicolumn{1}{c}{57}         \\
rank 2                 & 1           & 0           & 1           & 0           & \multicolumn{1}{c}{28}         \\
rank 3                 & 0           & 0           & 1           & 0           & \multicolumn{1}{c}{21}         \\
rank 4                 & 0           & 0           & 1           & 1           & \multicolumn{1}{c}{17}         \\
rank 5                 & 0           & 0           & 1           & 0           & \multicolumn{1}{c}{6}          \\
ranks $\ge 6$ & 46          & 16          & 128         & 44          & \multicolumn{1}{c}{32}         \\
\hline
\end{tabular}
\end{table}

\begin{table}[h]
%\caption{MetaCyc GAN-4}
\caption{Distribution of the true-compound rank in GAN-4 database searches against MetaCyc across benchmark spectral datasets}
\label{tab:MetaCyc-GAN-4}
\begin{tabular}{lccccl}
\hline
                       & CASMI2016SP & CASMI2017SP & CASMI2016FP & CASMI2017FP & \multicolumn{1}{c}{CASMI2022P} \\
\hline
total                  & 122         & 42          & 381         & 146         & \multicolumn{1}{c}{161}        \\
rank 1                 & 82          & 24          & 276         & 98          & \multicolumn{1}{c}{56}         \\
rank 2                 & 1           & 0           & 1           & 0           & \multicolumn{1}{c}{28}         \\
rank 3                 & 0           & 0           & 0           & 0           & \multicolumn{1}{c}{21}         \\
rank 4                 & 0           & 0           & 1           & 1           & \multicolumn{1}{c}{18}         \\
rank 5                 & 0           & 0           & 2           & 0           & \multicolumn{1}{c}{7}          \\
ranks $\ge 6$ & 39          & 18          & 101         & 47          & \multicolumn{1}{c}{31}         \\
\hline          
\end{tabular}
\end{table}

\begin{table}[h]
%\caption{MetaCyc GAN-5}
\caption{Distribution of the true-compound rank in GAN-5 database searches against MetaCyc across benchmark spectral datasets}
\label{tab:MetaCyc-GAN-5}
\begin{tabular}{lccccl}
\hline
                       & CASMI2016SP & CASMI2017SP & CASMI2016FP & CASMI2017FP & \multicolumn{1}{c}{CASMI2022P} \\
\hline
total                  & 122         & 42          & 381         & 146         & \multicolumn{1}{c}{161}        \\
rank 1                 & 91          & 35          & 308         & 107         & \multicolumn{1}{c}{68}         \\
rank 2                 & 0           & 0           & 1           & 0           & \multicolumn{1}{c}{28}         \\
rank 3                 & 0           & 0           & 0           & 1           & \multicolumn{1}{c}{25}         \\
rank 4                 & 0           & 0           & 0           & 0           & \multicolumn{1}{c}{5}          \\
rank 5                 & 0           & 0           & 0           & 1           & \multicolumn{1}{c}{7}          \\
ranks $\ge 6$ & 31          & 7           & 72          & 37          & \multicolumn{1}{c}{28}         \\
\hline          
\end{tabular}
\end{table}

\begin{table}[h]
%\caption{MetaCyc GAN-6}
\caption{Distribution of the true-compound rank in GAN-6 database searches against MetaCyc across benchmark spectral datasets}
\label{tab:MetaCyc-GAN-6}
\begin{tabular}{lccccl}
\hline
                       & CASMI2016SP & CASMI2017SP & CASMI2016FP & CASMI2017FP & \multicolumn{1}{c}{CASMI2022P} \\
\hline
total                  & 122         & 42          & 381         & 146         & \multicolumn{1}{c}{161}        \\
rank 1                 & 91          & 29          & 313         & 111         & \multicolumn{1}{c}{56}         \\
rank 2                 & 0           & 0           & 3           & 1           & \multicolumn{1}{c}{27}         \\
rank 3                 & 1           & 0           & 2           & 0           & \multicolumn{1}{c}{22}         \\
rank 4                 & 1           & 0           & 1           & 0           & \multicolumn{1}{c}{13}         \\
rank 5                 & 1           & 0           & 2           & 1           & \multicolumn{1}{c}{12}         \\
ranks $\ge 6$ & 28          & 13          & 60          & 33          & \multicolumn{1}{c}{31}         \\
\hline          
\end{tabular}
\end{table}

\begin{table}[h]
%\caption{MetaCyc GAN-7}
\caption{Distribution of the true-compound rank in GAN-7 database searches against MetaCyc across benchmark spectral datasets}
\label{tab:MetaCyc-GAN-7}
\begin{tabular}{lccccl}
\hline
                       & CASMI2016SP & CASMI2017SP & CASMI2016FP & CASMI2017FP & \multicolumn{1}{c}{CASMI2022P} \\
\hline
total                  & 122         & 42          & 381         & 146         & \multicolumn{1}{c}{161}        \\
rank 1                 & 99          & 37          & 327         & 119         & \multicolumn{1}{c}{60}         \\
rank 2                 & 0           & 0           & 0           & 0           & \multicolumn{1}{c}{24}         \\
rank 3                 & 0           & 0           & 0           & 0           & \multicolumn{1}{c}{25}         \\
rank 4                 & 0           & 0           & 0           & 0           & \multicolumn{1}{c}{6}          \\
rank 5                 & 0           & 0           & 0           & 0           & \multicolumn{1}{c}{12}         \\
ranks $\ge 6$ & 23          & 5           & 54          & 27          & \multicolumn{1}{c}{34}         \\
\hline          
\end{tabular}
\end{table}

\begin{table}[h]
%\caption{MetaCyc GAN-8}
\caption{Distribution of the true-compound rank in GAN-8 database searches against MetaCyc across benchmark spectral datasets}
\label{tab:MetaCyc-GAN-8}
\begin{tabular}{lccccl}
\hline
                       & CASMI2016SP & CASMI2017SP & CASMI2016FP & CASMI2017FP & \multicolumn{1}{c}{CASMI2022P} \\
\hline
total                  & 122         & 42          & 381         & 146         & \multicolumn{1}{c}{161}        \\
rank 1                 & 91          & 34          & 312         & 117         & \multicolumn{1}{c}{58}         \\
rank 2                 & 0           & 0           & 0           & 2           & \multicolumn{1}{c}{24}         \\
rank 3                 & 2           & 0           & 6           & 0           & \multicolumn{1}{c}{23}         \\
rank 4                 & 1           & 0           & 0           & 0           & \multicolumn{1}{c}{9}          \\
rank 5                 & 2           & 0           & 2           & 0           & \multicolumn{1}{c}{1}          \\
ranks $\ge 6$ & 26          & 8           & 61          & 27          & \multicolumn{1}{c}{46}         \\
\hline
\end{tabular}
\end{table}

\begin{table}[h]
%\caption{PubChem GAN-0}
\caption{Distribution of the true-compound rank in GAN-0 database searches against isomer decoys across benchmark spectral datasets (GAN-0 is trained using structural isomers from PubChem as negative samples)}
\label{tab:PubChem-GAN-0}
\begin{tabular}{lccccl}
\hline
                       & CASMI2016SP & CASMI2017SP & CASMI2016FP & CASMI2017FP & \multicolumn{1}{c}{CASMI2022P} \\
\hline
total                  & 122         & 42          & 381         & 146         & \multicolumn{1}{c}{161}        \\
rank 1                 & 109         & 32          & 354         & 146         & \multicolumn{1}{c}{32}         \\
rank 2                 & 1           & 1           & 5           & 0           & \multicolumn{1}{c}{11}         \\
rank 3                 & 1           & 1           & 3           & 0           & \multicolumn{1}{c}{8}          \\
rank 4                 & 0           & 1           & 2           & 0           & \multicolumn{1}{c}{6}          \\
rank 5                 & 2           & 1           & 2           & 0           & \multicolumn{1}{c}{5}          \\
ranks $\ge 6$ & 9           & 6           & 15          & 0           & \multicolumn{1}{c}{99}         \\
\hline          
\end{tabular}
\end{table}

\begin{table}[h]
%\caption{PubChem GAN-1}
\caption{Distribution of the true-compound rank in GAN-1 database searches against isomer decoys across benchmark spectral datasets}
\label{tab:PubChem-GAN-1}
\begin{tabular}{lccccl}
\hline
                       & CASMI2016SP & CASMI2017SP & CASMI2016FP & CASMI2017FP & \multicolumn{1}{c}{CASMI2022P} \\
\hline
total                  & 122         & 42          & 381         & 146         & \multicolumn{1}{c}{161}        \\
rank 1                 & 68          & 7           & 229         & 66          & \multicolumn{1}{c}{55}         \\
rank 2                 & 4           & 1           & 14          & 7           & \multicolumn{1}{c}{4}          \\
rank 3                 & 3           & 4           & 1           & 6           & \multicolumn{1}{c}{5}          \\
rank 4                 & 3           & 2           & 1           & 5           & \multicolumn{1}{c}{5}          \\
rank 5                 & 3           & 0           & 6           & 1           & \multicolumn{1}{c}{3}          \\
ranks $\ge 6$ & 41          & 28          & 130         & 61          & \multicolumn{1}{c}{89}         \\
\hline           
\end{tabular}
\end{table}

\begin{table}[h]
%\caption{PubChem GAN-2}
\caption{Distribution of the true-compound rank in GAN-2 database searches against isomer decoys across benchmark spectral datasets}
\label{tab:PubChem-GAN-2}
\begin{tabular}{lccccl}
\hline
                       & CASMI2016SP & CASMI2017SP & CASMI2016FP & CASMI2017FP & \multicolumn{1}{c}{CASMI2022P} \\
\hline
total                  & 122         & 42          & 381         & 146         & \multicolumn{1}{c}{161}        \\
rank 1                 & 70          & 8           & 240         & 67          & \multicolumn{1}{c}{36}         \\
rank 2                 & 3           & 1           & 6           & 7           & \multicolumn{1}{c}{1}          \\
rank 3                 & 2           & 1           & 6           & 1           & \multicolumn{1}{c}{10}         \\
rank 4                 & 1           & 1           & 5           & 3           & \multicolumn{1}{c}{6}          \\
rank 5                 & 0           & 0           & 6           & 0           & \multicolumn{1}{c}{4}          \\
ranks $\ge 6$ & 46          & 31          & 118         & 68          & \multicolumn{1}{c}{104}        \\
\hline           
\end{tabular}
\end{table}

\begin{table}[h]
%\caption{PubChem GAN-3}
\caption{Distribution of the true-compound rank in GAN-3 database searches against isomer decoys across benchmark spectral datasets}
\label{tab:PubChem-GAN-3}
\begin{tabular}{lccccl}
\hline
                       & CASMI2016SP & CASMI2017SP & CASMI2016FP & CASMI2017FP & \multicolumn{1}{c}{CASMI2022P} \\
\hline
total                  & 122         & 42          & 381         & 146         & \multicolumn{1}{c}{161}        \\
rank 1                 & 84          & 27          & 256         & 106         & \multicolumn{1}{c}{44}         \\
rank 2                 & 1           & 0           & 7           & 6           & \multicolumn{1}{c}{11}         \\
rank 3                 & 1           & 1           & 3           & 2           & \multicolumn{1}{c}{5}          \\
rank 4                 & 1           & 0           & 4           & 1           & \multicolumn{1}{c}{7}          \\
rank 5                 & 2           & 0           & 7           & 0           & \multicolumn{1}{c}{6}          \\
ranks $\ge 6$ & 33          & 14          & 104         & 31          & \multicolumn{1}{c}{88}         \\
\hline           
\end{tabular}
\end{table}

\begin{table}[h]
%\caption{PubChem GAN-4}
\caption{Distribution of the true-compound rank in GAN-4 database searches against isomer decoys across benchmark spectral datasets}
\label{tab:PubChem-GAN-4}
\begin{tabular}{lccccl}
\hline
                       & CASMI2016SP & CASMI2017SP & CASMI2016FP & CASMI2017FP & \multicolumn{1}{c}{CASMI2022P} \\
\hline
total                  & 122         & 42          & 381         & 146         & \multicolumn{1}{c}{161}        \\
rank 1                 & 94          & 25          & 300         & 112         & \multicolumn{1}{c}{52}         \\
rank 2                 & 3           & 2           & 9           & 2           & \multicolumn{1}{c}{16}         \\
rank 3                 & 2           & 0           & 4           & 1           & \multicolumn{1}{c}{10}         \\
rank 4                 & 1           & 0           & 6           & 3           & \multicolumn{1}{c}{5}          \\
rank 5                 & 2           & 1           & 8           & 1           & \multicolumn{1}{c}{6}          \\
ranks $\ge 6$ & 20          & 14          & 54          & 27          & \multicolumn{1}{c}{72}         \\
\hline           
\end{tabular}
\end{table}

\begin{table}[h]
%\caption{PubChem GAN-5}
\caption{Distribution of the true-compound rank in GAN-5 database searches against isomer decoys across benchmark spectral datasets}
\label{tab:PubChem GAN-5}
\begin{tabular}{lccccl}
\hline
                       & CASMI2016SP & CASMI2017SP & CASMI2016FP & CASMI2017FP & \multicolumn{1}{c}{CASMI2022P} \\
\hline
total                  & 122         & 42          & 381         & 146         & \multicolumn{1}{c}{161}        \\
rank 1                 & 96          & 35          & 312         & 114         & \multicolumn{1}{c}{50}         \\
rank 2                 & 3           & 2           & 3           & 3           & \multicolumn{1}{c}{8}          \\
rank 3                 & 0           & 0           & 1           & 3           & \multicolumn{1}{c}{5}          \\
rank 4                 & 1           & 0           & 3           & 0           & \multicolumn{1}{c}{10}         \\
rank 5                 & 1           & 0           & 2           & 2           & \multicolumn{1}{c}{4}          \\
ranks $\ge 6$ & 21          & 5           & 60          & 24          & \multicolumn{1}{c}{84}         \\
\hline           
\end{tabular}
\end{table}

\begin{table}[h]
%\caption{PubChem GAN-6}
\caption{Distribution of the true-compound rank in GAN-6 database searches against isomer decoys across benchmark spectral datasets}
\label{tab:PubChem GAN-6}
\begin{tabular}{lccccl}
\hline
                       & CASMI2016SP & CASMI2017SP & CASMI2016FP & CASMI2017FP & \multicolumn{1}{c}{CASMI2022P} \\
\hline
total                  & 122         & 42          & 381         & 146         & \multicolumn{1}{c}{161}        \\
rank 1                 & 114         & 40          & 313         & 135         & \multicolumn{1}{c}{58}         \\
rank 2                 & 0           & 0           & 2           & 2           & \multicolumn{1}{c}{16}         \\
rank 3                 & 0           & 0           & 2           & 1           & \multicolumn{1}{c}{4}          \\
rank 4                 & 0           & 0           & 2           & 0           & \multicolumn{1}{c}{6}          \\
rank 5                 & 1           & 0           & 4           & 0           & \multicolumn{1}{c}{8}          \\
ranks $\ge 6$ & 7           & 2           & 58          & 8           & \multicolumn{1}{c}{69}         \\
\hline           
\end{tabular}
\end{table}

\begin{table}[h]
%\caption{PubChem GAN-7}
\caption{Distribution of the true-compound rank in GAN-7 database searches against isomer decoys across benchmark spectral datasets}
\label{tab:PubChem-GAN-7}
\begin{tabular}{lccccl}
\hline
                       & CASMI2016SP & CASMI2017SP & CASMI2016FP & CASMI2017FP & \multicolumn{1}{c}{CASMI2022P} \\
\hline
total                  & 122         & 42          & 381         & 146         & \multicolumn{1}{c}{161}        \\
rank 1                 & 105         & 39          & 336         & 125         & \multicolumn{1}{c}{63}         \\
rank 2                 & 0           & 0           & 0           & 1           & \multicolumn{1}{c}{15}         \\
rank 3                 & 0           & 0           & 0           & 2           & \multicolumn{1}{c}{12}         \\
rank 4                 & 0           & 0           & 0           & 0           & \multicolumn{1}{c}{7}          \\
rank 5                 & 0           & 0           & 0           & 0           & \multicolumn{1}{c}{2}          \\
ranks $\ge 6$ & 17          & 3           & 45          & 18          & \multicolumn{1}{c}{62}         \\
\hline           
\end{tabular}
\end{table}

\begin{table}[]
%\caption{PubChem GAN-8}
\caption{Distribution of the true-compound rank in GAN-8 database searches against isomer decoys across benchmark spectral datasets}
\label{tab:PubChem-GAN-8}
\begin{tabular}{lccccl}
\hline
                       & CASMI2016SP & CASMI2017SP & CASMI2016FP & CASMI2017FP & \multicolumn{1}{c}{CASMI2022P} \\
\hline
total                  & 122         & 42          & 381         & 146         & \multicolumn{1}{c}{161}        \\
rank 1                 & 103         & 36          & 333         & 129         & \multicolumn{1}{c}{60}         \\
rank 2                 & 2           & 1           & 4           & 4           & \multicolumn{1}{c}{17}         \\
rank 3                 & 1           & 0           & 2           & 1           & \multicolumn{1}{c}{8}          \\
rank 4                 & 0           & 0           & 0           & 0           & \multicolumn{1}{c}{5}          \\
rank 5                 & 1           & 0           & 4           & 0           & \multicolumn{1}{c}{4}          \\
ranks $\ge 6$ & 15          & 5           & 38          & 12          & \multicolumn{1}{c}{67}         \\
\hline           
\end{tabular}
\end{table}

\clearpage
%\section{Supplementary References}
\bibliography{sn-bibliography}